\documentclass[lettersize,journal]{IEEEtran}
\usepackage{amsmath,amsfonts}
\usepackage{algorithmic}
\usepackage{algorithm}
\usepackage{array}
\usepackage[caption=false,font=normalsize,labelfont=sf,textfont=sf]{subfig}
\usepackage{hyperref}
\usepackage{textcomp}
\usepackage{stfloats}
\usepackage{url}
\usepackage{verbatim}
\usepackage{graphicx}
\usepackage{makecell}
\usepackage{booktabs}
\usepackage{cite}
\usepackage{multirow}
\usepackage{array}
\usepackage{xcolor}
\usepackage{cuted}   
\usepackage{capt-of} 
\usepackage{placeins}

\usepackage[most]{tcolorbox}
\usepackage{bm}
\usepackage{amsmath}
\usepackage{amsthm}
\usepackage{amsfonts}

\DeclareMathAlphabet{\mathpzc}{OT1}{pzc}{m}{it}

\DeclareFontFamily{U}{jkpmia}{}
\DeclareFontShape{U}{jkpmia}{m}{it}{<->s*jkpmia}{}
\DeclareFontShape{U}{jkpmia}{bx}{it}{<->s*jkpbmia}{}
\DeclareMathAlphabet{\mathfrak}{U}{jkpmia}{m}{it}

\definecolor{black}{rgb}{0.3,0.3,0.3}
\definecolor{blackb}{rgb}{0.3,0.3,0.3}
\definecolor{brownb}{rgb}{0.3,0.3,0.3}

\newtcbtheorem[no counter]{Definition}{Definition}{
  enhanced,
  rounded corners,
  attach boxed title to top left={
    yshifttext=-1mm
  },
  colback=white,
  colframe=black,
  fonttitle=\bfseries,
  coltitle=white,
  boxrule=0.5pt,
  boxed title style={
    rounded corners,
    size=small,
    colback=black,
    colframe=black,
  } 
}{def}

\makeatletter
\def\ps@IEEEtitlepagestyle{%
  \def\@oddhead{}\def\@evenhead{}%
  \def\@oddfoot{}\def\@evenfoot{}}
\makeatother
\begin{document}

\title{CWI: \textbf{C}omposite Humanoid \textbf{W}hole-Body \textbf{I}mitation System for Loco-manipulation}


\author{
Wenqi Ge$^{1,2,*}$,
Junde Guo$^{1, 3,*}$,
Zhen Fu$^{^{\dagger} 1, 3}$,
Shunpeng Yang$^{1,4}$,
Jiayu Chen$^{2}$,
Hua Chen$^{1,5}$
\thanks{$^{*}$ Equal contribution.}
\thanks{$^{\dagger}$ Project Lead.}
\thanks{$^{1}$ LimX Dynamics, Shenzhen 518055, China.}
\thanks{$^{2}$ The University of Hong Kong, Hong Kong SAR, China.}
\thanks{$^{3}$ School of Automation and Intelligent Manufacturing, Southern University of Science and Technology, Shenzhen 518055, China.}
\thanks{$^{4}$ Hong Kong University of Science and Technology, Hong Kong SAR, China.}
\thanks{$^{5}$ ZJU-UIUC Institute, Zhejiang University, Zhejiang 310058, China.}
}

\maketitle


\begin{strip}
\vspace{-9.5em}
    \centering
    \includegraphics[width=\textwidth]{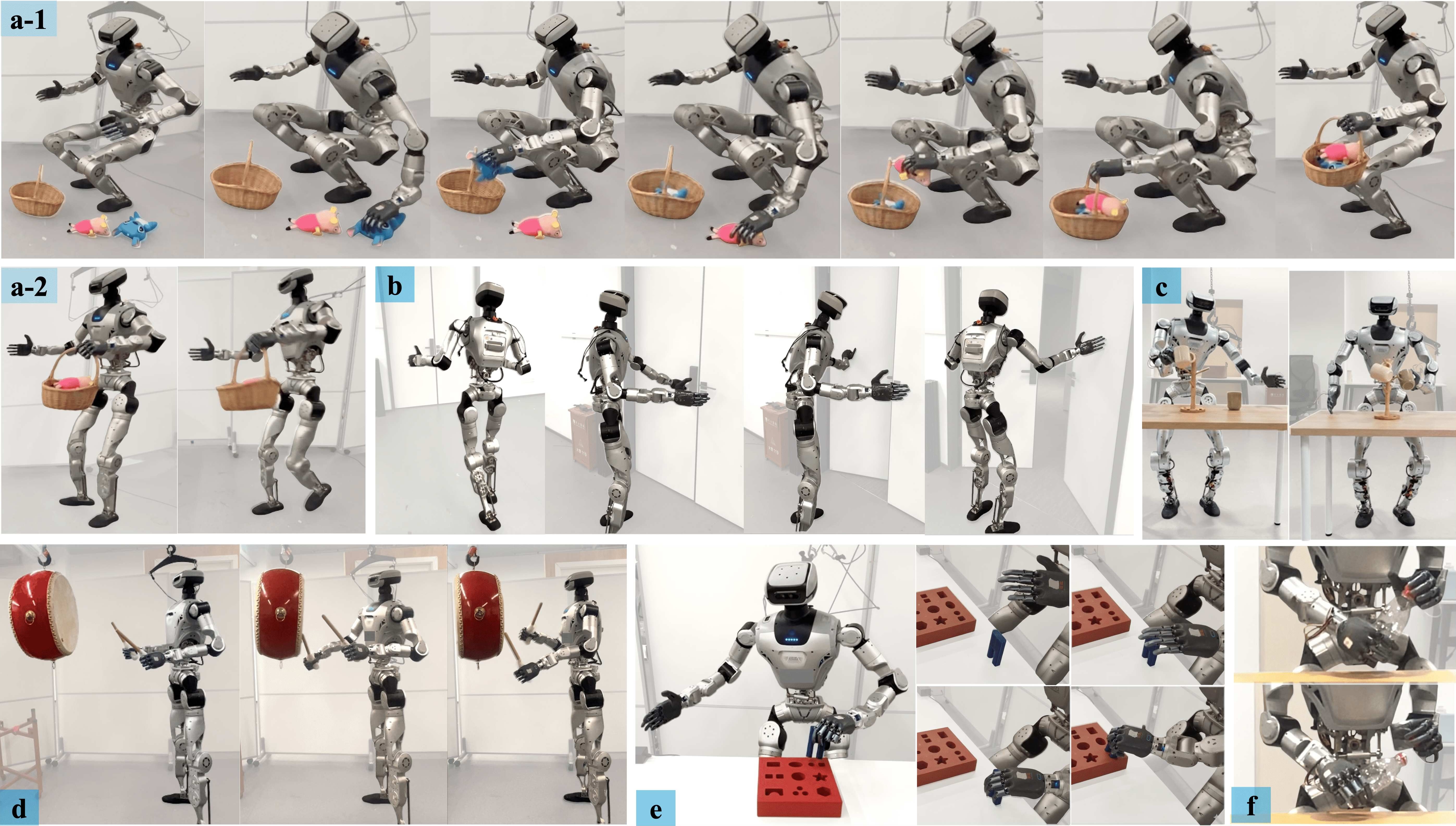}
    \captionof{figure}{CWI enables diverse whole-body loco-manipulation skills with stable locomotion and dexterous upper-body control. (a-1) Squatting to pick up toys and place them into a basket; (a-2) standing up and walking away while carrying the basket; (b) opening the door; (c) hanging cups on a rack; (d) striking a suspended drum; (e) precision assembly of a small part; (f) twisting a bottle cap.}
    \label{fig:placeholder}
\end{strip}

\FloatBarrier
\begin{abstract}
Achieving everyday tasks with humanoid robots requires coordinating stable locomotion with versatile manipulation. However, existing whole-body controllers still face significant challenges. Methods trained solely via command sampling, without motion-capture (MoCap) data, often struggle with sparse rewards and require carefully tuned curricula to converge. This is especially problematic for upper-body control, where the resulting motions deviate from human-like statistics and degrade whole-body coordination. Conversely, approaches that imitate full-body MoCap data suffer from dataset imbalance, as many locomotion trajectories are overly aggressive for stable-locomotion scenarios, necessitating extensive data filtering and augmentation. To address this, we present Composite Whole-Body Imitation (CWI), a framework that decouples the use of MoCap data for upper-body manipulation and lower-body locomotion. This decoupling allows us to exploit the full MoCap dataset of diverse manipulation references, while stable, command-conditioned lower-body locomotion is guided by dual discriminators trained on curated expert-quality walking and squatting clips via an Adversarial Motion Prior (AMP). A multi-critic architecture reduces conflicts among locomotion, manipulation, and motion-style objectives, and a teacher--student distillation stage yields a whole-body policy conditioned only on bimanual hand poses and velocity/height commands. We evaluate CWI through simulation experiments and real-world deployment on a full-size LimX Oli humanoid. The results show competitive loco-manipulation performance, robust whole-body coordination, and practical teleoperation without full-body motion-capture equipment. A project page with supplementary material can be found at \href{https://cwi-ral.github.io/CWI-RAL-Webpage/}{cwi-ral.github.io/CWI-RAL-Webpage}.
\end{abstract}


\section{Introduction}
\IEEEPARstart{H}{umanoid} robots are expected to operate in everyday human environments and eventually perform a wide range of household and industrial tasks \cite{gu2025humanoid}. Achieving these tasks requires coordinated loco-manipulation on humanoids \cite{wang2024autonomous}. With recent advances in reinforcement learning (RL) and imitation learning, remarkable progress has been made in locomotion \cite{sun2025learning,shi2025adversariallocomotionmotionimitation} and manipulation \cite{fu2024humanplus,lin2025sim} for humanoids. However, robust loco-manipulation remains a significant challenge, as stable locomotion and diverse upper-body manipulation often impose conflicting objectives on high-DoF systems.

Given the human-like morphology of humanoids, learning from human MoCap data has become a natural choice, and recent work increasingly integrates imitation learning with RL to exploit human motion priors \cite{liao2025beyondmimic,zhang2025track,he2025hover,he2024omnih2o,zhang2025falcon}. Parallel trends push toward more general whole-body control, either by scaling motion tracking to foundation-policy regimes \cite{luo2025sonic} or by mixing heterogeneous data sources to jointly improve agility and stability \cite{pan2025agility}, while promptable behavioral models trained via unsupervised RL \cite{li2025bfm} aim to cover diverse skills without task-specific supervision. Despite this progress, full-body MoCap datasets are often distributionally imbalanced: upper-body trajectories are rich and versatile, aligning well with manipulation, while lower-body trajectories frequently contain unstable behaviors and lack sufficient coverage of velocity and height ranges. This imbalance is especially acute for contact-rich loco-manipulation and human-object interaction, where both reliable base stability and compliance are required \cite{lu2025gentlehumanoid,wei2025hmc}. This motivates leveraging expert-quality lower-body data to stabilize locomotion while preserving the expressiveness of upper-body motions for manipulation.

Existing loco-manipulation controllers largely follow two paradigms: \textbf{Unified whole-body tracking} \cite{ji2024exbody2,li2025clone,sun2025ulc,ze2025twistteleoperatedwholebodyimitation}, which can achieve high fidelity but relies on extensive data curation and MoCap infrastructure, limiting deployment scenarios; \textbf{Decoupled frameworks} \cite{cheng2024expressive,ben2025homie,li2025amo,zhang2025falcon}, which separate manipulation and locomotion control through strategies such as IK-based upper-body tracking \cite{li2025amo}, separate upper/lower-body policies \cite{zhang2025falcon}, or joint-level command sampling for upper-body control \cite{wei2025hmc,lu2025gentlehumanoid}. However, many of these methods learn lower-body behaviors mainly through reward shaping, without sufficiently exploiting human lower-body motion priors.

To address this issue, we propose \textbf{Composite Whole-Body Imitation (CWI)}, a framework that decouples motion-data usage for upper-body manipulation and lower-body locomotion. The upper body is trained with the full unfiltered AMASS corpus to capture diverse upper-body references. The lower body is guided by a small set of curated walking and squatting clips through Adversarial Motion Priors (AMP), providing stable command-conditioned locomotion. Our contributions are as follows:

\begin{itemize}
    \item \textbf{Data Usage Decoupling for Loco-Manipulation.} Instead of discarding useful upper-body motions when MoCap trajectories are filtered for lower-body stability, we propose a decoupled MoCap data usage strategy that exploits the full AMASS upper-body corpus for diverse and smooth manipulation behavior. At the same time, curated walking and squatting clips are used to provide stable and natural lower-body behavior priors. This replaces full-corpus lower-body tracking with a curated small set, reducing the reference-data burden while preserving human-motion supervision in training.

    \item \textbf{Unified Whole-Body Training Framework with Decoupled Objectives.} Although CWI decouples motion data and rewards by role, it trains a unified whole-body policy instead of deploying separate upper- and lower-body controllers. Dual AMP discriminators provide walking and squatting style priors, while a multi-critic architecture assigns separate value heads to locomotion tracking, upper-body tracking, and style rewards, reducing optimization interference among these objectives. A teacher--student distillation stage then distills the full-reference teacher into a student policy controlled by bimanual hand poses and velocity/height commands, avoiding deployment-time full-body MoCap tracking.

    \item \textbf{Comprehensive Evaluation on a Full-Size Humanoid.} We compare CWI with representative baselines in simulation and validate it on a full-size humanoid robot across diverse real-world loco-manipulation tasks. The deployed system uses a portable Meta Quest VR headset and hand controllers for real-time loco-manipulation teleoperation, demonstrating practical operation without complex full-body motion-capture equipment.

\end{itemize}

\section{Background and Related Work}

\subsection{Learning-Based Humanoid Whole-Body Control}\label{sec:rel-a}
Learning-based methods have recently achieved strong locomotion robustness on legged systems. For humanoids, whole-body imitation tracking from filtered MoCap datasets can produce real--world deployable performance on in-distribution trajectories \cite{he2024omnih2o}, but tends to be sensitive to out-of-distribution (OOD) trajectories and often requires substantial data curation \cite{ji2024exbody2,pan2025agility}. Refinements such as the teacher--student framework \cite{ze2025twistteleoperatedwholebodyimitation}, and the closed-loop error correction with data augmentation \cite{li2025clone} to improve robustness on perturbed trajectories. However, these methods still require the lower body to imitate MoCap trajectories frame by frame. As a result, their lower-body behavior coverage remains bounded by the filtered dataset's distribution and does not naturally provide stable control of velocity and height for deployment.

A complementary line decouples locomotion and manipulation by conditioning on locomotion commands (velocity and height) and upper-body pose. \cite{ben2025homie} trains the legs with RL while sampling upper-body joints, and \cite{sun2025ulc} designs a curriculum from locomotion to manipulation, but the sampled upper-body trajectories can diverge from human motion and yield less coordinated behavior at deployment. Hierarchical designs such as \cite{li2025amo} blend trajectory optimization with RL, augmenting torso data and optimizing a stable standing pose for each upper-body configuration; when tracking velocity commands, however, the lower-body reference remains near standing, limiting gait expressivity and whole-body coherence. \cite{xue2025hugwbc} unifies versatile locomotion modes through gait-mode switching, but its focus is on locomotion diversity rather than coordinated manipulation.

Unlike these decoupled controllers, CWI decouples only MoCap data usage rather than the policy architecture.  It preserves upper-body coordination and stable command-conditioned lower-body control within a unified whole-body policy.

\subsection{Motion Imitation in Humanoid Robots}\label{sec:rel-b}

RL-based motion imitation, popularized by DeepMimic~\cite{peng2018deepmimic}, has since evolved into adversarial variants such as AMP~\cite{peng2021amp,xu2023adaptnet,xu2023composite}, which trains a discriminator as a task-agnostic motion prior over motion clips. AMP-style imitation has transferred well to real robots for quadrupedal locomotion~\cite{espinal2016quadrupedal,chen2025between} and humanoid locomotion-only controllers~\cite{zhang2024whole,lin2025hwc}, while DeepMimic-style whole-body tracking on humanoids~\cite{he2024omnih2o} typically relies on carefully filtered, task-specific reference datasets. However, these imitation paradigms are still limited for deployable loco-manipulation. 
AMP-based humanoid methods mainly learn reusable locomotion style priors, but have not been used to preserve diverse upper-body manipulation behaviors. 
Tracking-based methods can reproduce whole-body references, but their supervision is tied to specific reference trajectories rather than a reusable lower-body style prior. 
This makes it difficult to combine stable command-conditioned locomotion with diverse upper-body manipulation within a unified policy.

In contrast, CWI adopts an objective-specific imitation design rather than applying the same imitation objective to the whole body. 
For lower-body behavior, dual AMP discriminators learn stable motion priors from curated locomotion clips. For upper-body tracking, the full AMASS upper-body corpus is used to preserve diverse upper-body behavior. This decoupling makes better data usage, enabling stable command-conditioned locomotion and diverse upper-body behaviors within a unified policy.

\section{Method}
\subsection{Reinforcement Learning Problem Formulation} \label{sec:Meth-1}
We model the task as a POMDP and learn a policy $\pi_{\theta}: \Omega \rightarrow \mathcal{A}$ that maximizes the expected discounted return. A teacher--student framework is used for robust deployment: the teacher policy $\pi^{\text{tea.}}$ operates on the full state $\mathbf{s}_t$, while the deployable student $\pi^{\text{stu.}}$ receives a history of proprioceptive observations $\mathbf{o}_{t-H:t}^{\text{stu.}}$. Both policies output an action $\mathbf{a}_t \in \mathbb{R}^{31}$; the control command $\mathbf{q}_t^* = \mathbf{q}_0 + \alpha \mathbf{a}_t$ is tracked by a Proportional--Derivative (PD) controller, where $\mathbf{q}_0$ is the nominal pose and $\alpha$ the action scale.

\begin{figure*}[!htbp]
    \centering
    \includegraphics[width=\textwidth]{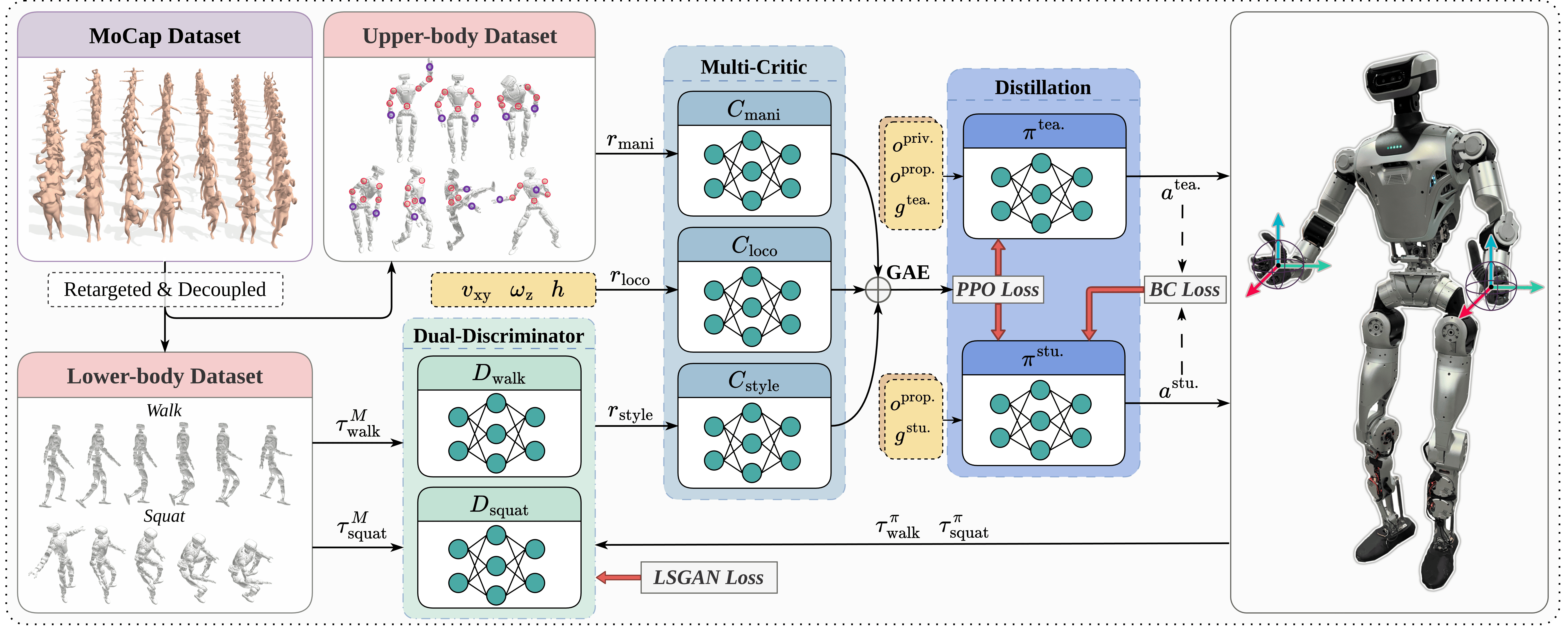}
    \caption{Overview of the Composite Whole-Body Imitation (CWI) framework. Our method utilizes large-scale upper-body data for manipulation with curated lower-body motions for stable locomotion. Dual discriminators guide the policy to learn lower-body behaviors (Walk/Squat) from compact datasets, while the upper body tracks diverse AMASS trajectories. The unified policy is trained with a multi-critic architecture and then distilled into a student policy controlled by velocity/height commands and bimanual hand poses.}
    \label{fig:framework}
\end{figure*}

\subsection{Data Decoupling for Loco-Manipulation} \label{sec:Meth-2}
\label{sec:decoupling}

\begin{figure}[t]
    \centering
    \includegraphics[width=\linewidth]{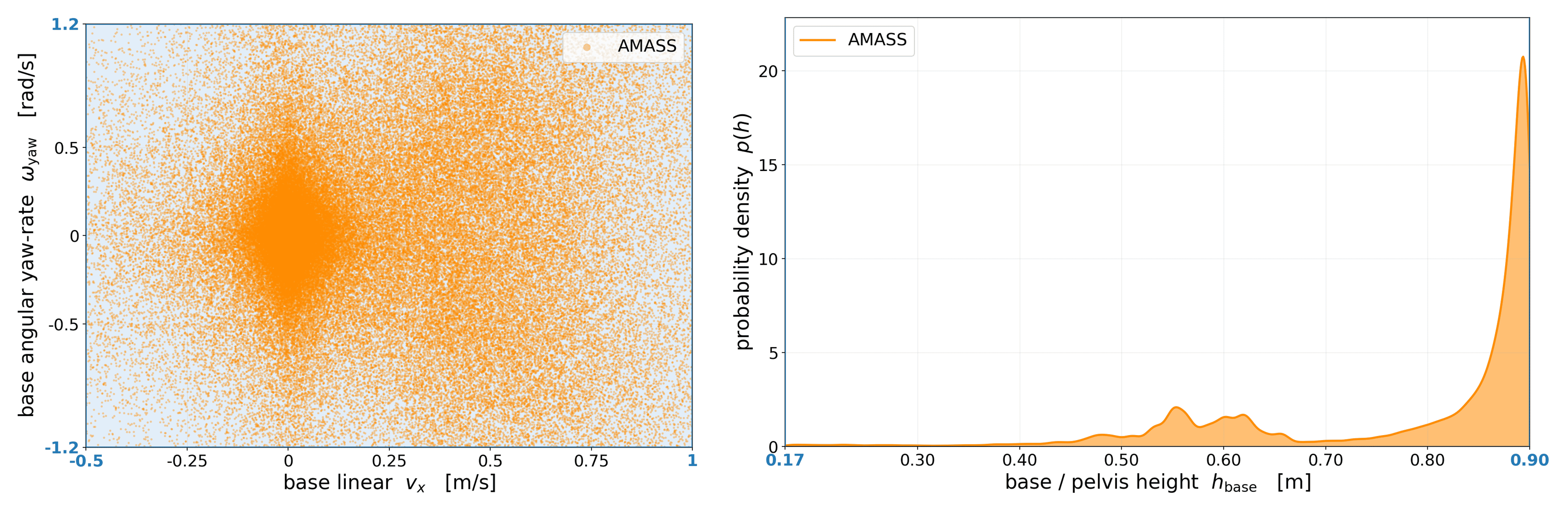}
    \caption{AMASS lower-body distribution vs.\ our training command range (blue). \emph{Left}: pelvis-height density peaks near $0.85$--$0.90$\,m, leaving the low-height end of $h\!\in\![0.17,0.9]$\,m sparse. \emph{Right}: base $(v_x,\omega_{\text{yaw}})$ scatter clusters near the origin, leaving the boundaries of the commanded box ($v_x\!\in\![-0.5,1.0]$\,m/s, $\omega_{\text{yaw}}\!\in\![-1.2,1.2]$\,rad/s) sparsely populated. This imbalance motivates curating a small lower-body library $\mathcal{M}_l$ instead of using raw AMASS lower-body trajectories.}
    \label{fig:amass-dist}
\end{figure}

Large-scale MoCap datasets offer rich motion priors but are often biased toward highly dynamic motions incompatible with stable loco-manipulation. Filtering these out wastes valuable upper-body data. Furthermore, the remaining locomotion data is often stylistically inconsistent and unbalanced across velocities and heights, making it unsuitable for robust, command-conditioned control. Fig.~\ref{fig:amass-dist} illustrates this imbalance: the AMASS pelvis-height density concentrates near upright standing ($\sim$0.85--0.90\,m), and the base linear/angular velocities cluster around the origin, leaving the low-height and faster-walking corners of our command box severely under-represented. We therefore retain the full AMASS upper-body \cite{mahmood2019amass} corpus for versatile manipulation tracking in the robot's base frame, but curate a small, well-supervised set of walking and squatting clips for the lower-body library $\mathcal{M}_l$, with one AMP discriminator per class (Sec.~\ref{sec:Meth-3}) since the two style classes are qualitatively distinct.

\textbf{Decoupled Motion Dataset.} We retarget the source MoCap data to our humanoid using a kinematic optimizer similar to PHC~\cite{Luo2023PerpetualHC}. This process optimizes key body positions to match human poses while explicitly regularizing temporal smoothness to avoid jitter and ensure continuous, robot-feasible motion. From the retargeted corpus, we construct the upper-body datasets $\mathcal{M}_{\text{u}}$ containing all AMASS upper-body trajectories, and a lower-body dataset $\mathcal{M}_{\text{l}}$ comprising approximately ten curated expert-quality clips per category (omnidirectional walking and squatting).

All motions in $\mathcal{M}_{\text{u}}$ are transformed relative to the robot's base frame, isolating the arm workspace from locomotion. This re-expression also decouples upper-body trainability from the source clip's root motion: even dynamic AMASS clips (e.g., fast spins, jumps) supply clean, human-feasible arm trajectories once the root is removed, so we keep the full AMASS upper-body corpus without filtering. The upper-body reference combines joint- and link-level targets:
\begin{equation}
\label{eq1}
    m_{\text{u}} = \{ \mathbf{q}_{\text{u}}^{\text{ref}}, \mathbf{p}_k^{\text{ref}}, \mathbf{R}_k^{\text{ref}}, \tilde{\mathbf{p}}_k^{\text{ref}} \}_{k \in \mathcal{K}_{\text{u}}}
\end{equation}
where $\mathbf{q}_{\text{u}}^{\text{ref}} \in \mathbb{R}^{17}$ are upper-body joint positions covering the waist (yaw/roll/pitch) and arms, and for each link $k \in \mathcal{K}_{\text{u}}=\{\text{shoulder, elbow, wrist, hand}\}$, $\mathbf{p}_k^{\text{ref}} \in \mathbb{R}^3$, $\mathbf{R}_k^{\text{ref}}$, and $\tilde{\mathbf{p}}_k^{\text{ref}} \in \mathbb{R}^9$ denote the position, orientation, and keypoint-based pose~\cite{Allshire_2022} respectively. Placing the waist in the upper-body group lets torso bending be supervised jointly with arm motion by the same AMASS clip. Because AMASS contains very few head motions, the neck DoFs and the head link are excluded from $m_{\text{u}}$; instead, the head command $\mathbf{q}_{\text{h}}^{\text{cmd}} \in \mathbb{R}^2$ covering neck yaw and pitch is sampled directly within its mechanical range at the start of each episode and supervised by a separate quadratic tracking term, so the head can be driven independently from a VR headset at deployment.

The lower-body reference for AMP training encapsulates base dynamics, articulation, and task-space features:
\begin{equation}
    m_{\text{l}} = \{ \mathbf{v}_{\text{xy}}^{\text{ref}}, \boldsymbol{\omega}_{\text{z}}^{\text{ref}}, h^{\text{ref}}, \mathbf{q}_{\text{l}}^{\text{ref}}, \dot{\mathbf{q}}_{\text{l}}^{\text{ref}},\mathbf{p}_k^{\text{ref}}, \mathbf{R}_k^{\text{ref}} \} _{k \in \mathcal{K}_{\text{l}}}
\end{equation}
where $\mathbf{q}_{\text{l}}^{\text{ref}}, \dot{\mathbf{q}}_{\text{l}}^{\text{ref}} \in \mathbb{R}^{12}$ are lower-body joint positions and velocities, and $k \in \mathcal{K}_{\text{l}}=\{\text{hip, knee, ankle}\}$.

\textbf{Decoupled Goal Interfaces.}
The locomotion goal is $g_l = [v^{\mathrm{cmd}}_{xy}, \omega^{\mathrm{cmd}}_z, h^{\mathrm{cmd}}] \in \mathbb{R}^4$,
which we decompose into a velocity component
$g_l^{\mathrm{vel}} = [v^{\mathrm{cmd}}_{xy}, \omega^{\mathrm{cmd}}_z]$
and a height component
$g_l^{h} = [h^{\mathrm{cmd}}]$
for discriminator selection (Sec.~\ref{sec:Meth-3}).
Although $\mathcal{M}_{\text{l}}$ is narrow along the style axis (only walking and squatting clips), $g_l$ is sampled uniformly across the full deployment command range during training. The policy learns to respond to all operator-commanded velocities and heights, while the curated clips provide lower-body style priors rather than fixed tracking references. For the upper body, the teacher receives the full reference $g^{\text{tea}}_{\text{u}} = m_{\text{u}}$, while the student uses only dual hand poses for teleoperation:

\begin{equation}
    g^{\text{stu}}_{\text{u}} = [ \tilde{\mathbf{p}}_{\text{L}}^{\text{ref}}, \tilde{\mathbf{p}}_{\text{R}}^{\text{ref}} ] \in \mathbb{R}^{18}
\end{equation}
where $\tilde{\mathbf{p}}_{\text{L}}^{\text{ref}}$ and $\tilde{\mathbf{p}}_{\text{R}}^{\text{ref}}$ are the 9D keypoint features of the left and right hands as defined in Eq. \ref{eq1}.

\subsection{Composite Imitation Learning Objective} \label{sec:Meth-3}
\textbf{Dual discriminators.} To produce stable locomotion under varying commands while accommodating upper-body motion, we build on AMP and introduce dual discriminators that capture style priors for walking and squatting from the curated clips in $\mathcal{M}_l$. Each discriminator $D^k_\phi$ ($k\!\in\!\{\text{walk},\text{squat}\}$) scores a lower-body state-transition window $\tau_t = s^{\text{lower}}_{t-3:t}$ (four frames of joint and link states, excluding upper-body signals so the prior is purely lower-body) and is trained with a least-squares GAN objective with gradient penalty:
\begin{equation}
\begin{aligned}
\underset{\phi}{\arg \min } & \ \mathbb{E}_{\tau\sim\mathcal{M}_k}\left[\left(D_{\phi}^k\left(\tau\right)-1\right)^{2}\right]  +\mathbb{E}_{\tau \sim d^{\pi}}\left[\left(D^k_{\phi}\left(\tau\right)+1\right)^{2}\right] \\
& +\frac{w^{\mathrm{gp}}}{2} \mathbb{E}_{\tau\sim\mathcal{M}_k}\left[\left\|\left.\nabla_{\tau} D^k_{\phi}(\tau)\right|\right\|^{2}\right]
\end{aligned}
\end{equation}
where $w^{\text{gp}}$ stabilizes training. The active discriminator is selected by the locomotion command: $D_{\text{walk}}$ when $g_l^{\text{vel}}\!\neq\!0$ and $D_{\text{squat}}$ otherwise (height tracking). It supplies a continuous style reward that rewards outputs classified as expert-like:
\begin{equation}
r^{\mathrm{style}} =
\begin{cases}
\max\!\left(0,\; 1 - 0.25\bigl(D^{\mathrm{walk}}_\phi(\tau) - 1\bigr)^2\right), & \text{if } g_l^{\mathrm{vel}} \neq 0, \\
\max\!\left(0,\; 1 - 0.25\bigl(D^{\mathrm{squat}}_\phi(\tau) - 1\bigr)^2\right), & \text{if } g_l^{\mathrm{vel}} = 0.
\end{cases}
\end{equation}

\textbf{Task rewards.} The total reward is partitioned into three groups (Tab.~\ref{tab:reward}): the style reward $r^{\text{style}}$ above; a locomotion task reward $r^{\text{loco}}$ tracking $g_l$ via exponential terms; and an upper-body tracking reward $r^{\text{upper}}$ with exponential $L_2$ losses on $\boldsymbol{q}_u^{\text{ref}}$ and the triangle-keypoint targets $\tilde{\boldsymbol{p}}_u^{\text{ref}}$~\cite{Allshire_2022} from $\mathcal{M}_u$. Standard regularizers (projected-gravity alignment, action smoothness, torque penalties) are assigned to the group they primarily affect.

\begin{table}  
    \centering
    \caption{Task and style rewards.}
    \label{tab:reward}
    \begin{tabular}{lcl}
        \hline
        Loco. Reward $r^{\text {loco}}$          & Definition                         & Weights \\ \hline
        Linear Velocity       & $\exp \!\big(-(v_{xy}-v_{xy}^{\text{cmd}})^2 \big)$          & 3.0     \\
        Angular Velocity      & $\exp\!\big(-(\omega_z-{\omega_{z}^{\text{cmd}}})^2 \big)$      & 2.5     \\ 
        Height tracking       & $\exp\!\big(-(h-h^{\text{cmd}})^{2} \big)$   & 2.5    \\  \hline \hline
        Manip. Reward $r^{\text {upper}}$ & Definition                         & Weights \\ \hline\noalign{\vskip 1pt}
        DoF Position & $\exp\!\big(-\|\boldsymbol{q}^{\text{upper}}-\boldsymbol{q}^{\text{upper}}_{\text{ref}}\|^{2} / \sigma_{q}^{2}\big)$ & 2.0 \\ \noalign{\vskip 1pt}
        Tri-Keypoint Position & $\exp\!\big(-\|\tilde{\boldsymbol{p}}^{\text{upper}}-\tilde{\boldsymbol{p}}^{\text{upper}}_{\text{ref}}\|^{2} / \sigma_{p}^{2}\big)$  & 3.0 \\ \noalign{\vskip 1pt}\hline \hline
        Style Reward $r^{\text {style}}$ & Definition                         & Weights \\ \hline
        AMP & $\max\!\bigl(0,\; 1 - 0.25(D^k_\phi(\tau) - 1)^2\bigr)$  & 2.0 \\ \hline 
    \end{tabular}
\end{table}

\textbf{Multi-Critic Actor Learning.} Sharing a single critic across locomotion, upper-body tracking, and style rewards creates optimization conflicts, exacerbated by the high-variance adversarial signal. We therefore assign each of the three groups $i\!\in\!\{\text{loco},\text{upper},\text{style}\}$ its own value function $V^i$, compute group-wise GAE advantages, and normalize within each minibatch:
\begin{equation}
\begin{aligned}
\delta_{t}^{i} &= r_{t}^{i} + \gamma V^{i}\!\left(o_{t+1}\right) - V^{i}\!\left(o_{t}\right), \\
\hat{A}_{t}^{i} &= \frac{A_{t}^{i} - \mu_{i}}{\sigma_{i} + \varepsilon},
\end{aligned}
\label{eq:adv_sum}
\end{equation}
where $A_t^i$ is the unnormalized GAE from $\delta_t^i$ and $\mu_i,\sigma_i$ are minibatch statistics. The overall advantage $\hat{A}_t = \sum_i w_i \hat{A}_t^i$ is plugged into the standard PPO clipped surrogate loss:
\begin{equation}
L(\theta)=\hat{\mathbb{E}}\left[\min \left(r_{t}(\theta) \hat{A}_t, \operatorname{clip}\left(r_{t}(\theta), 1-\epsilon, 1+\epsilon\right) \hat{A}_t\right)\right]
\end{equation}
with importance ratio $r_t(\theta)$ and clipping parameter $\epsilon$.

\FloatBarrier
\subsection{Policy Distillation} \label{sec:Meth-4}

Single-stage RL whole-body tracking tends to be jerky or unstable~\cite{vijayan2025multi}, so we adopt a teacher--student framework. The teacher is trained with privileged information and the full upper-body reference $g^{\text{tea}}_u = m_u$; the deployable student receives only proprioceptive observations and the reduced command $g^{\text{stu}}_u = [\tilde{\mathbf{p}}_L^{\text{ref}}, \tilde{\mathbf{p}}_R^{\text{ref}}]$ (the 9D hand keypoints extracted from $m_u$). Because pure imitation cannot bridge this gap, we combine PPO with an action-matching BC term:
\begin{equation}
L\left(\pi_{\mathrm{stu}}\right)=L_{\mathrm{RL}}\left(\pi_{\mathrm{stu}}\right)+\lambda \mathbb{E}_{o\sim\rho}\!\left[
\left\| \mu_{\text{stu}}(o^{\text{stu}})-\mu_{\text{tea}}(o^{\text{tea}})\right\|_{}^{2}
\right]
\end{equation}
where $L_{\text{RL}}$ uses the teacher's reward, $\rho$ is the state distribution from student rollouts (teacher queried on-policy), and $\lambda$ is annealed to balance imitation and exploration. This RL+BC distillation consistently outperforms pure BC and pure RL, yielding smoother motion and better generalization.

Reducing the student command to bimanual hand poses is a deployment-driven trade-off: operators typically supply only EEF targets, and online IK reconstruction would risk discontinuities near singularities. Distilling on hand poses sidesteps this and constrains the student to the AMASS manifold, at the cost of a small open-loop accuracy gap relative to the joint-command setting.

\FloatBarrier
\section{Experiments}
We evaluate CWI through three studies: (i) a quantitative benchmark against representative whole-body controllers; (ii) ablations isolating the contribution of AMP, multi-critic, distillation, and the AMASS upper-body prior; and (iii) real-world deployment on a LimX Oli humanoid driven by a commodity VR headset.
\FloatBarrier
\subsection{Experiment Setup}\label{sec:exp-1}
Evaluations use the 31-DoF LimX Oli ($1.65$\,m, $50$\,kg). Policies are trained in IsaacLab. Locomotion commands are sampled within $v_x\!\in\![-0.5,1.0]$\,m/s, $\omega_{\text{yaw}}\!\in\![-1.2,1.2]$\,rad/s, and $h\!\in\![0.17,0.9]$\,m. We use the full AMASS upper-body corpus, split into $85\%/15\%$ for training and evaluation, with upper-body references expressed in the robot's base frame. Standard domain randomization (mass, friction, sensor noise, latency) is applied for sim-to-real transfer.

\textbf{Interface.} All policies output $a_t \in \mathbb{R}^{31}$. CWI uses two student variants distilled from the same teacher with comparable end-effector accuracy: a \emph{comparison variant} that takes an extra joint input $g_u = [\mathbf{q}_u^{\text{ref}}, \tilde{\mathbf{p}}_L^{\text{ref}}, \tilde{\mathbf{p}}_R^{\text{ref}}]$ matching the joint-driven baselines' interface for fair benchmarking, and a \emph{deployment variant} $g_u = [\tilde{\mathbf{p}}_L^{\text{ref}}, \tilde{\mathbf{p}}_R^{\text{ref}}]$ matching the minimalist hand-pose signal naturally supplied by VR teleoperation devices, used in the ablation and hardware experiments.

\textbf{Metrics.} We report standard tracking metrics in the base-relative frame (lower is better unless noted): success rate \textbf{SR}~(\%, fraction of episodes balanced with mean per-keybody distance to reference below $0.5$\,m); per-keybody position and orientation errors $p^{\text{err}}$~(mm), $R^{\text{err}}$~(rad); and base velocity / height tracking errors $v^{\text{err}}$~(m/s), $\omega^{\text{err}}_{\text{yaw}}$~(rad/s), $h^{\text{err}}$~(mm). In the comparison study, $p^{\text{err}}/R^{\text{err}}$ aggregate over all upper-body keypoints; in the ablation, we instead report end-effector errors $p_{ee}^{\text{err}}/R_{ee}^{\text{err}}$ averaged over the two hands (joint-space metrics omitted, matching the deployment-style command).

\subsection{Comparison Study}
We compare CWI against several representative baselines. Since the robot platforms in the original works differ from ours, we re-implemented all baselines (denoted by $^{*}$) on the LimX Oli robot, with only their \emph{framework architectures} following the official codebases. The observation composition, reward terms with weights, and PD gains are unified across all methods, since the baselines' native settings differ substantially across papers; this keeps the comparison focused on framework design and data usage differences. Evaluation uses locomotion commands sampled from the defined ranges and upper-body motions from AMASS sequences unseen during training.

\begin{itemize}
  \item $\textbf{HOVER}^{*}$: A unified whole-body imitation setup in which a teacher policy with privileged full-body state tracks MoCap, distilled into a DAgger-style student that consumes a $25$-step proprioceptive history; we follow the original per-episode Bernoulli($0.5$) mode$\times$sparsity command masking applied separately to upper and lower body, with action-MSE as the distillation objective.
  \item $\textbf{FALCON}^{*}$: A single-stage decoupled controller with dual PPO agents for lower and upper-body control, jointly trained in a shared environment; both policies receive the full proprioceptive state and both goal vectors, paired with asymmetric privileged critics. Upper-body targets are retargeted from AMASS via IK. We omit FALCON’s force-perturbation curriculum because heavy-object robustness is outside our evaluation scope and is not used by the other compared methods.
  \item $\textbf{HOMIE}^{*}$: A decoupled controller where RL is used for lower-body control, and PD control is used for upper-body. We retain HOMIE's upper-body action-ratio curriculum with reward-triggered ramp-up, $1$\,s target resampling-with-interpolation, the knee--height-coupled squat reward with $1/3$ of environments forced into squat mode per resample, mirror augmentation paired with a symmetry loss on actor/critic outputs, and per-hand mass randomization.
  \item \textbf{CWI}: The proposed composite controller, as detailed in Sec.~\ref{sec:Meth-3}.
\end{itemize}

\begin{figure}[!htbp]
    \centering
    \includegraphics[width=0.5\textwidth]{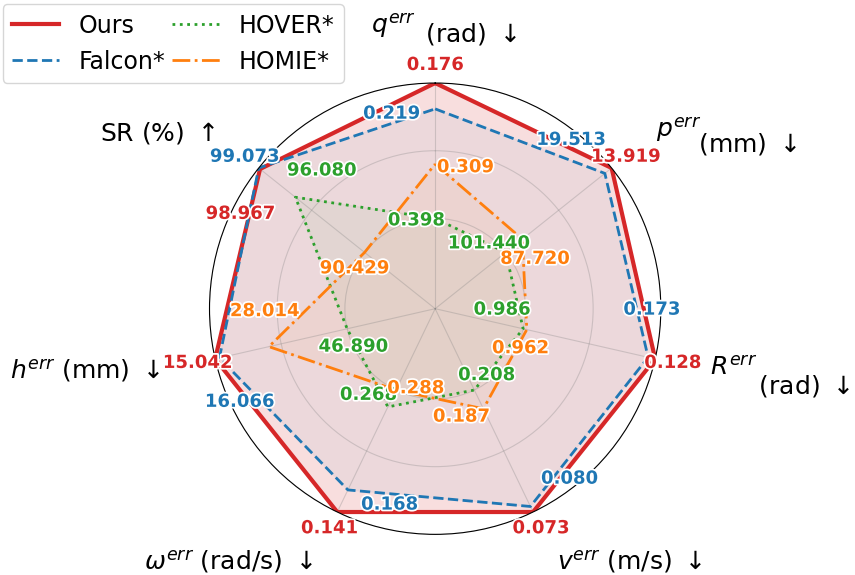}
    \caption{Quantitative comparison of loco-manipulation controllers in simulation. Each metric is min-max normalized across all methods so that larger area indicates better overall performance. Raw values are annotated on each axis. CWI achieves the best trade-off across nearly all metrics.}
    \label{fig:com-study}
\end{figure}

As shown in Fig.~\ref{fig:com-study}, CWI surpasses baselines across nearly all the reported metrics when trained on the complete AMASS dataset without motion filtering. For upper-body tracking, $\text{FALCON}^*$ shows relatively higher arm tracking errors, which is consistent with its separate upper/lower policy structure. $\text{HOVER}^*$ shows reduced robustness when upper-body motions are paired with base commands beyond its training distribution. $\text{HOMIE}^*$ presents increased joint-level deviations, likely due to the PD-controlled upper body when following non-smooth references. For locomotion, CWI achieves the lowest velocity and height errors. The only metric where it does not rank first is success rate: $\text{FALCON}^*$ is higher by less than 0.1\%, likely because our policy maintains strict upper-body tracking even under boundary commands, which can occasionally induce slight instability. Overall, CWI achieves a superior trade-off between precise upper-body tracking and stable, command-conditioned locomotion by decoupling objectives at both the data and learning levels.

\FloatBarrier
\subsection{Ablation Study}\label{sec:exp-3}
We conduct ablations to quantify the contribution of each key component of CWI. Each variant is trained under the same setting, and evaluated with the metrics in Sec.~\ref{sec:exp-1}. To quantify motion naturalness, we additionally report a relaxed Dynamic Time Warping (DTW) distance~\cite{wen_constrained_2025}, which relaxes the strict frame-to-frame alignment constraint in standard DTW, allowing robust comparison to reference clips under small temporal phase shifts. As summarized in Tab.~\ref{tab:ablation}, \textbf{Baseline} denotes the full CWI system. We then consider five ablated variants:
\begin{itemize}
  \item \textbf{w/o MC}: replacing the multi-critic architecture with a single critic;
  \item \textbf{w/o Distill}: removing the two-stage teacher--student scheme and instead trains a single-stage policy;
  \item \textbf{Single AMP}: pooling walking and squatting clips into a single AMP discriminator instead of one per class;
  \item \textbf{w/o AMP}: removing the AMP-based style reward and training without locomotion discriminators;
  \item \textbf{w/o AMASS-up}: replacing the AMASS upper-body reference with randomly sampled upper-body joint targets~\cite{ben2025homie}, holding all other components fixed.
\end{itemize}

\begin{figure}[!htbp]
    \centering
    \includegraphics[width=1.\linewidth]{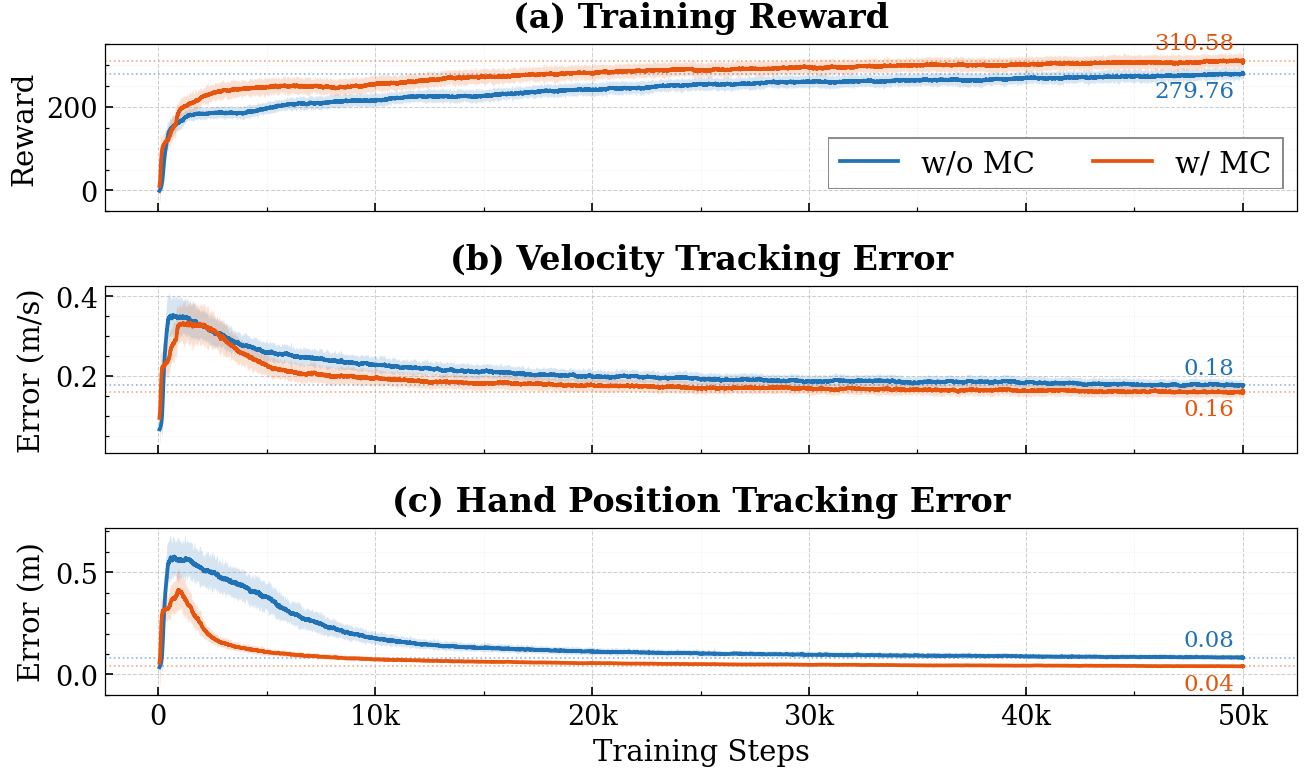}
    \caption{Training curves comparing single-critic (w/o mc) and multi-critic (w/ mc) variants. The multi-critic consistently achieves higher final reward and faster convergence in both velocity and hand-position errors.}
    \label{fig:ab-mcvel}
\end{figure}

\textbf{Effect of the multi-critic.} Replacing the multi-critic with a single critic leads to a consistent drop across both locomotion and manipulation metrics. As shown in Tab.~\ref{tab:ablation}, the single-critic variant exhibits higher angular velocity error and notably worse end-effector tracking $p_{\text{ee}}^{\text{err}}$, $R_{\text{ee}}^{\text{err}}$. The training curves in Fig.~\ref{fig:ab-mcvel} further confirm that the multi-critic variant converges faster and achieves higher final reward with lower velocity and hand-position errors, suggesting that a single shared critic suffers from interference between competing objectives.

\textbf{Effect of two-stage training.} Without distillation, the single-stage policy fails to learn upper-body manipulation from the sparse hand-pose command alone -- end-effector errors increase sharply ($p_{ee}^{\text{err}}\!:42.91\!\to\!173.2$\,mm, $R_{ee}^{\text{err}}\!:0.171\!\to\!0.672$\,rad, marked $^{\dagger}$ in Tab.~\ref{tab:ablation}); locomotion-only metrics are reported for transparency but are not directly comparable. The two-stage procedure is therefore critical: the teacher first learns reliable whole-body control with privileged signals and dense upper-body references, then distills this knowledge to the student, enabling accurate end-effector tracking from only bimanual hand poses.

\textbf{Effect of AMP style priors.} Removing AMP entirely (\textit{w/o AMP}) leaves locomotion command-tracking only mildly worse but collapses style: $d_{\text{dtw}}$ jumps from $0.452$ to $1.413$\,rad, with frequent stance violations and foot slips. Collapsing the dual discriminators into a single one (\textit{Single AMP}) is a milder failure mode of the same kind -- style still degrades ($d_{\text{dtw}}\!:0.452\!\to\!0.615$\,rad, $\ddot{q}_{p95}\!:28.8\!\to\!35.1$\,rad/s$^2$) and locomotion tracking shows slight regression ($v_{xy}^{\text{err}}\!:0.100\!\to\!0.110$\,m/s, $\omega_z^{\text{err}}\!:0.183\!\to\!0.192$\,rad/s) -- because mode-averaging across walking and squat priors blurs each. End-effector tracking is essentially unchanged in both variants ($p_{ee}^{\text{err}}\!\approx\!42.9$\,mm), confirming that the dual-discriminator design improves lower-body style without trading off manipulation accuracy.

\textbf{Effect of AMASS upper-body data.} Training the upper-body policy on randomly sampled joint targets~\cite{ben2025homie} rather than AMASS leaves it unprepared for the human-motion statistics at evaluation, where AMASS trajectories are used as the test reference. The upper body becomes out-of-distribution, especially for fast trajectories, and end-effector tracking degrades sharply ($p_{ee}^{\text{err}}\!:42.91\!\to\!62.32$\,mm, $R_{ee}^{\text{err}}\!:0.171\!\to\!0.275$\,rad, $\ddot{q}_{p95}\!:28.8\!\to\!48.2$\,rad/s$^2$). The resulting upper-body disturbance further destabilizes the lower body, degrading every locomotion metric ($v_{xy}^{\text{err}}\!:0.100\!\to\!0.131$\,m/s, $\omega_z^{\text{err}}\!:0.183\!\to\!0.249$\,rad/s, $h^{\text{err}}\!:19.65\!\to\!24.20$\,mm). This isolates the AMASS prior from the decoupled architecture: the architecture supplies the structural decoupling, but only the AMASS data supplies the human-aligned upper-body trajectories whose statistics the deployment task actually demands.

\begin{table}[t]
    \centering
    \caption{Ablation study results using LimX Oli}
    \label{tab:ablation}
    \resizebox{\columnwidth}{!}{%
        \renewcommand{\arraystretch}{1.3}
        \begin{tabular}{lccccccc}
            \toprule
            \textbf{Method} &
            \textbf{\begin{tabular}[c]{@{}c@{}}$v_{\text{xy}}^{\text{err}}$\\ (m/s)\end{tabular}} &
            \textbf{\begin{tabular}[c]{@{}c@{}}$\omega_{\text{z}}^{\text{err}}$\\ (rad/s)\end{tabular}} &
            \textbf{\begin{tabular}[c]{@{}c@{}}$h^{\text{err}}$\\ (mm)\end{tabular}} &
            \textbf{\begin{tabular}[c]{@{}c@{}}$p_{\text{ee}}^{\text{err}}$\\ (mm)\end{tabular}} &
            \textbf{\begin{tabular}[c]{@{}c@{}}$R_{\text{ee}}^{\text{err}}$\\ (rad)\end{tabular}} &
            \textbf{\begin{tabular}[c]{@{}c@{}}$d_{\text{dtw}}$\\ (rad)\end{tabular}} &
            \textbf{\begin{tabular}[c]{@{}c@{}}$\ddot{\boldsymbol{q}}_{p95}$\\ (rad/s$^2$)\end{tabular}} \\
            \midrule
            \textbf{Baseline}                 & 0.100             & \textbf{0.1825}   & \textbf{19.65}    & \textbf{42.91}                   & 0.1708                           & \textbf{0.452}    & \textbf{28.8} \\
            \textbf{w/o MC}                   & \textbf{0.099}    & 0.199             & 20.64             & 55.49                            & 0.2308                           & 0.520             & 30.4 \\
            \textbf{w/o Distill}$^{\dagger}$  & \textit{\textcolor{gray!60}{0.099}} & \textit{\textcolor{gray!60}{0.147}} & \textit{\textcolor{gray!60}{19.43}} & \textcolor{red}{173.2} & \textcolor{red}{0.6723} & \textit{\textcolor{gray!60}{0.444}} & \textit{\textcolor{gray!60}{28.1}} \\
            \textbf{Single AMP}               & 0.110             & 0.192             & 21.07             & 42.93                            & 0.1702                           & 0.615             & 35.1 \\
            \textbf{w/o AMP}                  & 0.125             & 0.242             & 22.52             & 42.92                            & \textbf{0.1696}                  & \textcolor{red}{1.413} & 39.3 \\
            \textbf{w/o AMASS-up}             & 0.131             & 0.249             & 24.20             & 62.32                            & 0.275                            & 0.508             & 48.2 \\
            \bottomrule
        \end{tabular}%
    }
    \vspace{2pt}
    \par\noindent{\footnotesize\raggedright
    \textbf{Bold}: best per column; \textcolor{red}{red}: severely degraded; $\ddot{\boldsymbol{q}}_{p95}$: 95th-percentile joint acceleration over all DoFs and the evaluation horizon.\par
    $^{\dagger}$\,\textit{w/o Distill} fails upper-body tracking; other metrics not ranked.\par}
\end{table}

\subsection{Real-World Experiments}

\textbf{Qualitative Evaluation.}
We evaluate CWI on diverse real-world loco-manipulation tasks, as shown in Fig.~\ref{fig:placeholder}. These include precision tasks such as bottle cap twisting (f) and small-part assembly (e), whole-body coordination tasks such as opening a door (b) and striking a suspended drum (d), and long-horizon tasks involving sequential picking, placing, and carrying (a). Across all scenarios, the system maintains stable locomotion while executing dexterous upper-body motions. For additional experiments and video demonstrations, please refer to our project website.

\textbf{Coordination Analysis.}
\begin{figure*}[!htbp]
    \centering
    \includegraphics[width=\linewidth]{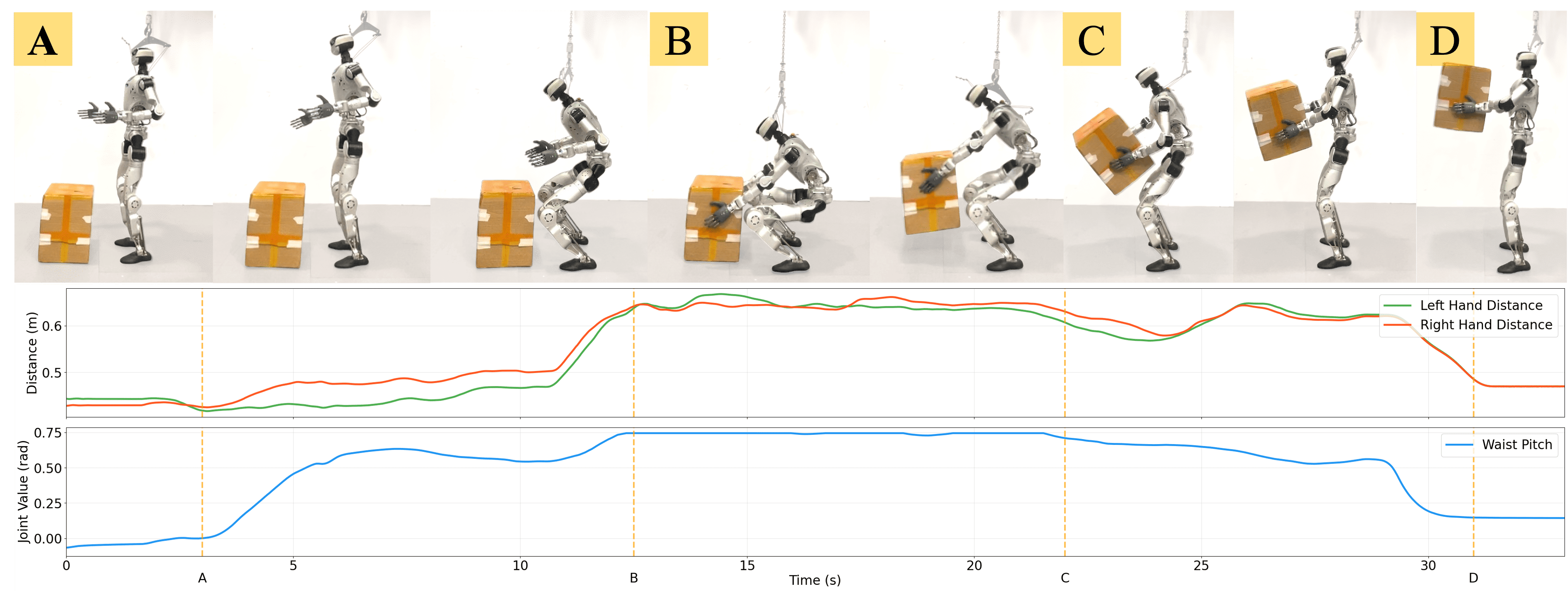}
    \caption{Real-world whole-body coordination during a box-lifting task. Top: snapshots of four phases — (A) approaching and squatting to reach the box, (B) grasping and beginning to lift, (C) raising the box with both hands, and (D) standing up while holding the box overhead. Bottom: left/right hand distances from the base and waist pitch joint value over time. The waist pitch automatically increases as the hands extend forward and downward, and decreases as the robot stands upright — without any explicit torso command.}
    \label{fig:waist-coord}
\end{figure*}
To further illustrate the learned whole-body coordination, we analyze a teleoperated box-lifting sequence. As shown in Fig.~\ref{fig:waist-coord}, when the operator commands the hands to reach forward and downward (phases A--B), the waist pitch automatically increases to extend the reachable workspace; as the hands retract and rise (phases C--D), the torso straightens accordingly. This coordinated behavior emerges purely from the learned policy without any explicit torso commands — the upper-body MoCap data in AMASS naturally captures the coupling between hand reach and torso posture, and our composite imitation objective preserves this coordination while maintaining stable locomotion throughout the task.

\section{Conclusion}
We presented CWI, a unified framework for humanoid loco-manipulation that decouples MoCap data usage by role through composite imitation. The full AMASS upper-body corpus provides diverse manipulation references, while compact curated walking and squatting clips provide lower-body motion priors through dual AMP-based discriminators. A multi-critic architecture improves optimization across locomotion, manipulation, and style objectives, while teacher--student distillation produces a whole-body policy controlled by bimanual hand poses and velocity/height commands. Experiments in simulation demonstrate competitive loco-manipulation performance. Real-world tests on a full-size LimX Oli humanoid further show robust whole-body coordination and portable teleoperation without full-body motion-capture equipment.

\textbf{Limitations and Future Work.} CWI is designed for portable loco-manipulation with a simple user interface, so its behavior coverage is limited by both the motion libraries and the command interface. Since the policy is controlled by bimanual hand poses and velocity/height commands, it does not support arbitrary joint-level or contact-level commands for the whole body, such as deliberately using a foot to press a trash-bin pedal or using an elbow as the primary contact point. Future work will use the portable teleoperation interface to collect multimodal demonstrations in everyday scenarios. The compact and task-oriented command space provides structured supervision for learning higher-level autonomous policies, including vision-language-action (VLA) models.

\bibliographystyle{IEEEtran}
\bibliography{ref}

\end{document}